\documentclass[10pt,twocolumn,letterpaper]{article}

\usepackage{cvpr}              

\usepackage{graphicx}
\usepackage{amsmath}
\usepackage{amssymb}
\usepackage{booktabs}

\usepackage[pagebackref,breaklinks,colorlinks]{hyperref}

\usepackage[capitalize]{cleveref}
\crefname{section}{Sec.}{Secs.}
\Crefname{section}{Section}{Sections}
\Crefname{table}{Table}{Tables}
\crefname{table}{Tab.}{Tabs.}
\usepackage{algorithm}
\usepackage{algpseudocode}
\usepackage{multirow}
\usepackage{graphicx}
\usepackage{caption}
\usepackage{wrapfig}
\usepackage{graphicx}
\usepackage{bbm}
\usepackage{bibentry}

\newtheorem{definition}{Definition}
\newtheorem{proposition}{Proposition}
\newtheorem{proof}{Proof}

\def\cvprPaperID{4175} 
\def\confName{CVPR}
\def\confYear{2023}
\newcommand\blfootnote[1]{%
  \begingroup
  \renewcommand\thefootnote{}\footnote{#1}%
  \addtocounter{footnote}{-1}%
  \endgroup
}

\begin{document}

\title{Deep Fair Clustering via Maximizing and Minimizing Mutual Information: Theory, Algorithm and Metric}
\author{Pengxin Zeng$^\ast$, Yunfan Li$^\ast$, Peng Hu, Dezhong Peng, Jiancheng Lv, Xi Peng$^\dagger$\\
College of Computer Science, Sichuan University, China\\
{\tt\small \{zengpengxin.gm, yunfanli.gm, penghu.ml, pengx.gm\}@gmail.com;}\\
{\tt\small \{pengdz, lvjiancheng\}@scu.edu.cn}
}

\maketitle
\blfootnote{$^\ast$ Equal contribution}
\blfootnote{$^\dagger$ Corresponding author}
\begin{abstract}
   Fair clustering aims to divide data into distinct clusters while preventing sensitive attributes (\textit{e.g.}, gender, race, RNA sequencing technique) from dominating the clustering. Although a number of works have been conducted and achieved huge success recently, most of them are heuristical, and there lacks a unified theory for algorithm design. In this work, we fill this blank by developing a mutual information theory for deep fair clustering and accordingly designing a novel algorithm, dubbed FCMI. In brief, through maximizing and minimizing mutual information, FCMI is designed to achieve four characteristics highly expected by deep fair clustering, \textit{i.e.}, compact, balanced, and fair clusters, as well as informative features. Besides the contributions to theory and algorithm, another contribution of this work is proposing a novel fair clustering metric built upon information theory as well. Unlike existing evaluation metrics, our metric measures the clustering quality and fairness as a whole instead of separate manner. To verify the effectiveness of the proposed FCMI, we conduct experiments on six benchmarks including a single-cell RNA-seq atlas compared with 11 state-of-the-art methods in terms of five metrics. The code could be accessed from \url{ https://pengxi.me}.
\end{abstract}

\section{Introduction}
\label{Sec:Introduction}
\begin{figure}
\centering
\includegraphics[width=\linewidth]{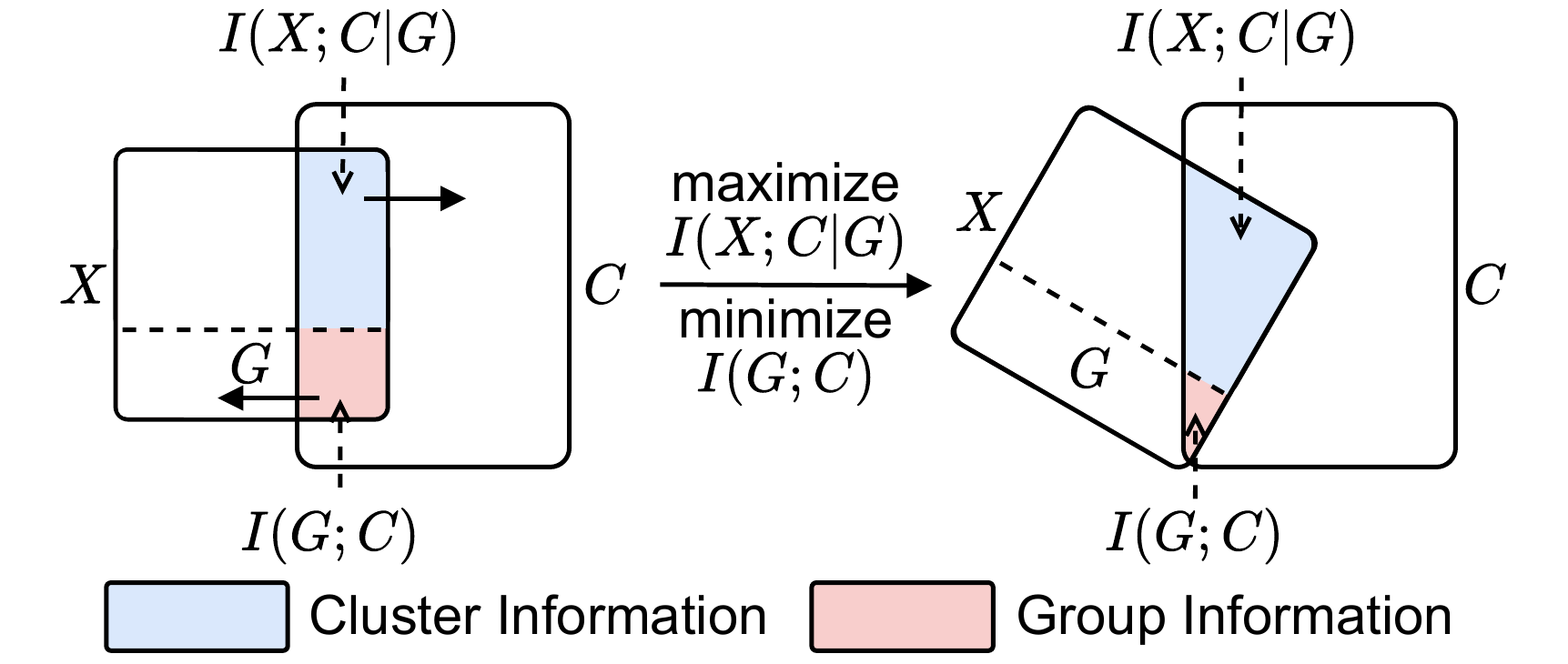}
\caption{Illustration of our basic idea using information diagrams, where $X$, $G$, and $C$ denote the inputs, sensitive attributes, and clustering assignments, respectively. To cluster data, we expect the overlap between non-group information and cluster information (\textit{i.e.}, the conditional mutual information $I(X; C|G)$) to be maximized. Meanwhile, to prevent sensitive attributes from dominating the clustering results, we expect the overlap between group information and cluster information (\textit{i.e.}, the mutual information $I(G, C)$) to be minimized.}
\label{Fig:Basic Idea}
\end{figure}
Clustering plays an important role in machine learning ~\cite{liu2012robust,liu2010robust,luo2021bi,shen2017compressed,yang2016ell,yi2013semi,kang2018unified}, which could partition data into different clusters without any label information. It has been widely used in many real-world applications such as multi-view learning~\cite{wang2018detecting,tao2017ensemble}, image segmentation~\cite{li2015temporal}, and bioinformatics~\cite{kiselev2019challenges}. In practice, however, the data might be confounded with sensitive attributes (\textit{e.g.}, gender, race, etc., also termed as \textbf{group information}) that probably overwhelm the intrinsic semantic of samples (also termed as \textbf{cluster information}). Taking single-cell RNA clustering as a showcase, standard methods would partition data based on sequencing techniques (group information) instead of intrinsic cell types (cluster information), since cells sequenced by different techniques would result in different expression levels~\cite{tran2020benchmark} and most clustering methods cannot distinguish these two kinds of information. The case is similar in many automatic learning systems where the clustering results are biased toward sensitive attributes, which would interfere with the decision-making~\cite{A_snapshot_of_the_frontiers_of_fairness_in_machine_learning,Empirical_risk_minimization_under_fairness_constraints,50_years_of_test__un__fairness__Lessons_for_machine_learning}. Notably, even though these sensitive attributes are known in prior, it is daunting to alleviate or even eliminate their influence, \textit{e.g.}, removing the ``gender'' information from the photos of users.

As a feasible solution, fair clustering aims to hide sensitive attributes from the clustering results. Commonly, a clustering result is considered fair when samples of different sensitive attributes are uniformly distributed in clusters so that the group information is protected. However, it would lead to a trivial solution if the fairness is over-emphasized, \textit{i.e.}, all samples are assigned to the same cluster. Hence, in addition to fairness, balance and compactness are also highly expected in fair clustering. Specifically, a balanced clustering could avoid the aforementioned trivial solution brought by over-emphasized fairness, and the compactness refers to a clear cluster boundary.

To achieve fair clustering, many studies have been conducted to explore how to incorporate fairness into lustering~\cite{fairlets,Scalable_fair_clustering,SpFC,FCC,FALG,Towards,DFCV,DFDC}. Their main differences lie in \textit{i)} the stage of fairness learning, and \textit{ii)} the depth of the model. In brief,~\cite{fairlets,Scalable_fair_clustering} incorporate the fairness in a pre-processing fashion by packing data points into so-called \textit{fairlets} with balanced demographic groups and then partitioning them with classic clustering algorithms. \cite{SpFC, FCC} are in-processing methods that formulate fairness as a constraint for clustering. As a representative of post-processing methods, \cite{FALG} first performs classic clustering and then transforms the clustering result into a fair one by linear programming. Different from the above shallow models, \cite{Towards,DFCV,DFDC} propose performing fair clustering in the latent space learned by different deep neural networks to boost performance. Although promising results have been achieved by these methods, almost all of them are heuristically and empirically designed, with few theoretical explanations and supports. In other words, it still lacks a unified theory to guide the algorithm design.

In this work, we unify the deep fair clustering task under the mutual information theory and propose a novel theoretical-grounded deep fair clustering method accordingly. As illustrated in Fig.~\ref{Fig:Basic Idea}, we theoretically show that clustering could be achieved by maximizing the conditional mutual information (CMI) $I(X; C|G)$ between inputs $X$ and cluster assignments $C$ given sensitive attributes $G$. Meanwhile, we prove that the fairness learning could be formulated as the minimization of the mutual information (MI) $I(G; C)$. In this case, sensitive attributes will be hidden in the cluster assignments and thus fair clustering could be achieved. To generalize our theory to deep neural networks, we additionally show a deep variant could be developed by maximizing the mutual information $I(X; X^\prime)$ between the input $X$ and its approximate posterior $X^\prime$. Notably, some deep clustering methods~\cite{IMSAT, Yijie} have been proposed based on the information theory. However, they are remarkably different from this work. To be exact, they ignored the group information. As a result, the group information will leak into cluster assignments, leading to unfair partitions. In addition, we prove that our mutual information objectives intrinsically correspond to four characteristics highly expected in deep fair clustering, namely, compact, balanced, and fair clusters, as well as informative features.

Besides the above contributions to theory and algorithm, this work also contributes to the performance evaluation. To be specific, we notice that almost all existing methods evaluate clustering quality and fairness separately. However, as fair clustering methods usually make a trade-off between these two aspects, such an evaluation protocol might be partial and inaccurate. As an improvement, we design a new evaluation metric based on the information theory, which simultaneously measures the clustering quality and fairness. The contribution of this work could be summarized as follows:
\begin{itemize}
    \item We formulate deep fair clustering as a unified mutual information optimization problem. Specifically, we theoretically show that fair clustering could be achieved by maximizing CMI between inputs and cluster assignments given sensitive attributes while minimizing MI between sensitive attributes and cluster assignments. Moreover, the informative feature extraction could be achieved by maximizing MI between the input and its approximate posterior.
    \item Driven by our unified mutual information theory, we propose a deep fair clustering method and carry out extensive experiments to show its superiority on six fair clustering benchmarks, including a single-cell RNA atlas.
    \item To evaluate the performance of fair clustering more comprehensively, we design a novel metric that measures the clustering quality and fairness as a whole from the perspective of information theory.
\end{itemize}

\section{Related Work}
\label{Sec.Relative Work}
To alleviate or even eliminate the influence of sensitive attributes, many efforts have been devoted on fair clustering~\cite{Fair_k_center_clustering_for_data_summarization, Clustering_without_over_representation, Proportionally_fair_clustering, Making_existing_clusterings_fairer, Individual_fairness_for_k_clustering, A_pairwise_fair_and_community_preserving_approach_to_k_center_clustering}. Based on how fairness is incorporated, the existing works could be roughly divided into three categories, namely, pre-processing, in-processing, and post-processing methods. In brief, the pre-processing methods endow classic clustering methods with hand-craft fairness constraints. For example, Chierichetti et al.~\cite{fairlets} first divide data points into several subsets (\textit{i.e.}, \textit{fairlets}) with the fairness constraint, and then employ a classic clustering algorithm on these \textit{fairlets} to obtain the data partition. However, the fairlets construction requires at least quadratic running time, which is daunting in practice. To improve the scalability, Backurs et al.~\cite{Scalable_fair_clustering} employ a tree metric to approximate the fairlets construction in nearly linear time. Different from the pre-processing methods, the in-processing methods recast fairness as a constraint for joint optimization with the clustering objective. For example, Kleindessner et al.~\cite{SpFC} recast the fairness as a linear constraint and embed it into the spectral clustering. Ziko et al.~\cite{FCC} propose a variational framework by integrating fairness as a Kullback-Leibler (KL) term into the classic clustering methods. Opposite from the pre-processing methods, the post-processing methods~\cite{FALG} aim to transform the given clustering result into a fair one by solving a linear programming problem.

Motivated by the success of deep clustering~\cite{AE,yang2019deep,li2021contrastive,guo2017deep,ghasedi2017deep}, some studies have been carried out on deep fair clustering. For example, Wang et al.~\cite{Towards} propose learning a fair embedding by forcing the cluster centers to be equidistant from group centers, which could handle an arbitrary number of sensitive attributes. Li et al.~\cite{DFCV} make a step forward to explore fair clustering on large-scale and high-dimensional visual data by incorporating fairness through adversarial training. Very recently, Zhang et al.~\cite{DFDC} generate fair pseudo cluster assignments to guide the model optimization.

Although promising results have been achieved by these methods, their success partially relies on some tricks like pre-clustering and data augmentation which are clumsy in practical use. Besides, most existing works are designed heuristically and empirically, with few theoretical explanations and supports. Different from these studies, the proposed FCMI is built upon information theory, of which the working mechanism is interpretable. We reveal that deep fair clustering could be achieved by maximizing \textit{i)} the mutual information $I(X; X^\prime)$ between the input $X$ and its approximate posterior $X^\prime$, and \textit{ii)} the conditional mutual information $I(X; C|G)$ between $X$ and cluster assignments $C$ given sensitive attributes $G$, while minimizing the mutual information $I(G; C)$. Both theoretical analysis and experimental results demonstrate the effectiveness of our method.

\section{Method}
\label{Sec: Method}
In this section, we first give the mathematical definition of fair clustering. After that, we elaborate on how to learn compact, balanced, and fair clusters, as well as informative features through a unified information theory. Finally, we summarize the implementation of the proposed algorithm.
\begin{figure*}
  \centering
  \includegraphics[width=0.9\linewidth]{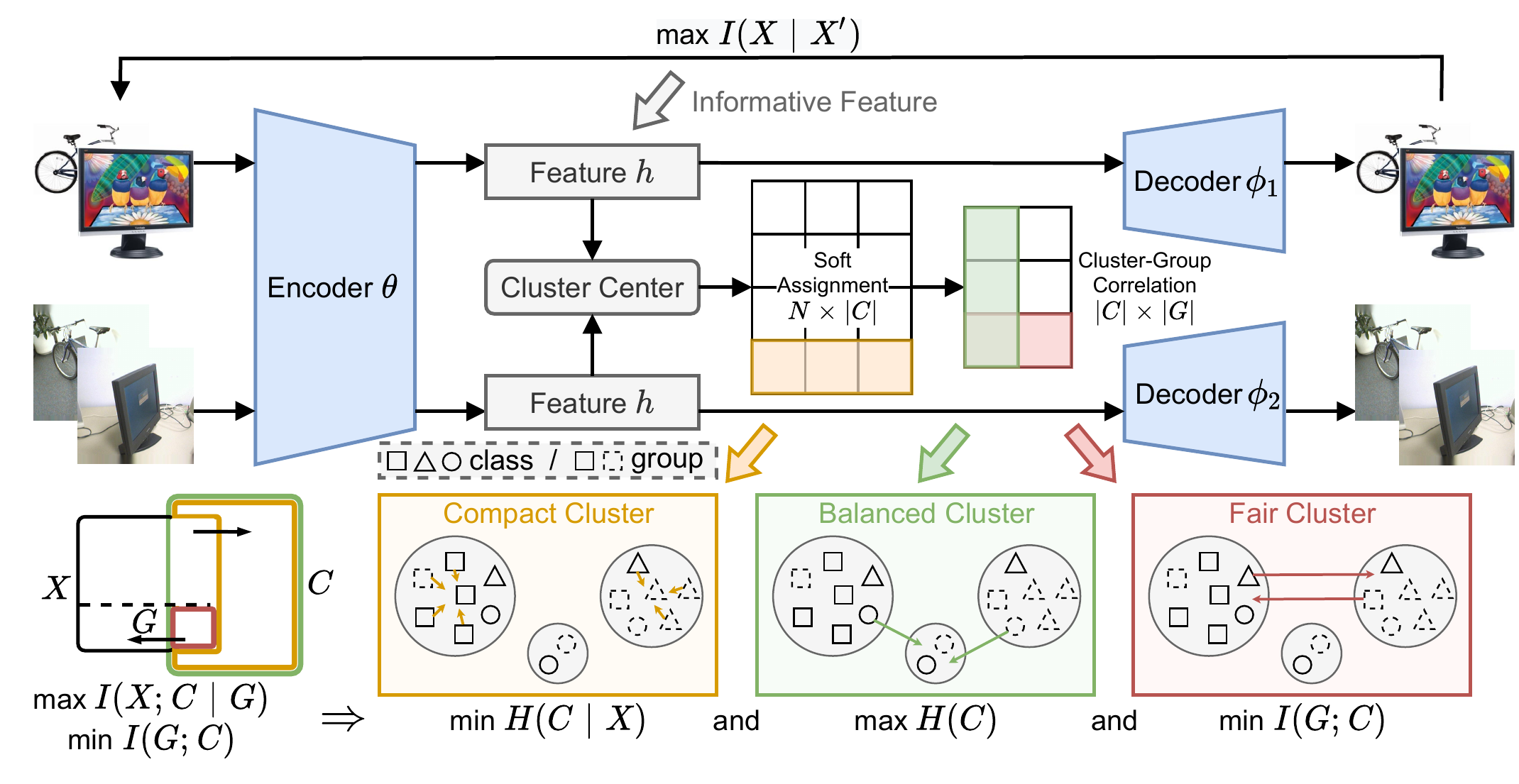}
  \vspace{-0.2cm}
  \caption{Overview of the proposed FCMI. In brief, FCMI extracts informative features by maximizing the mutual information $I(X; X^\prime)$ between samples $X$ and the corresponding posterior $X^\prime$, implementing as an auto-encoder. To learn compact and balanced clusters in the hidden space, we maximize the conditional mutual information $I(X; C|G)$ between $X$ and the cluster assignments $C$ given sensitive attributes $G$. Meanwhile, we minimize the mutual information $I(G; C)$ to endow the clusters with robustness against sensitive attributes. }
  \label{Fig:2}
\end{figure*}
\subsection{Problem Definition}
\label{Sec: Problem Definition}
For a given dataset $X = \{x_{1}, x_{2}, \dots, x_{N}\}$ with sensitive attributes $G = \{g_{1}, g_{2}, \dots, g_{N}\}$, fair clustering aims to partition $X$ into $K$ disjoint clusters with cluster assignments $C = \{c_{1}, c_{2}, \dots, c_{N}\}$ by alleviating or even eliminating the influence of $G$, where $g_{i} \in [1, 2, \dots, T]$, $c_{i} \in [1, 2, \dots, K]$, $g_{i} = j$ means that sample $x_i$ belongs to the $j$-\textit{th} group, $N$ is the data size, and $T$ is the group number. As pointed out by~\cite{DFCV}, a clustering result is considered absolutely fair if the cluster assignments only depend on the semantics and are independent of the sensitive attributes. Based on such a heuristic definition, we mathematically formulate fair clustering as follows.
\begin{definition}
\label{Def: fair}
\textbf{Fair Clustering.} Let $\tilde{g}_t=\{x_i|g_i=t\}$ and $\tilde{c}_k=\{x_i|c_i=k\}$ be the set of samples belonging to group $t$ and cluster $k$ respectively, and $p_{{\tilde{g}_t},{\tilde{c}_k}} = \frac{1}{N}{|\tilde{g}_{t} \cap \tilde{c}_{k}|}$ be the joint probability density of groups and clusters, the clustering result is absolutely fair if
\begin{equation}
    p_{{\tilde{g}_t},{\tilde{c}_k}} = p_{\tilde{g}_t}p_{\tilde{c}_k}, \forall t, k
\end{equation}
where $p_{\tilde{g}_t} = \frac{1}{N}{|\tilde{g}_{t}|}$ and $p_{\tilde{c}_k} = \frac{1}{N}{|\tilde{c}_{k}|}$ denote the marginal probability densities of groups and clusters respectively.
\end{definition}

Notably, fairness could be trivially achieved by assigning all samples to the same cluster, but obviously, it is not a reasonable solution. Hence, in addition to fairness, a good fair clustering method is also expected to embrace compactness and balance. Specifically, a clustering result is compact if the within-cluster distance is much smaller than the between-cluster distance, and a balanced clustering could avoid the aforementioned trivial solution due to over-emphasized fairness. For a deep fair clustering method, apart from the above three clustering characteristics, it also aims to learn informative features with redundancy removal for better clustering performance.

\subsection{Deep Fair Clustering via Maximizing and Minimizing Mutual Information}
\label{Sec: 3.2 MI}
As mentioned above, deep fair clustering has four objectives, namely, compact, balanced, fair clusters, and informative features, which are daunting to formulate and optimize jointly. In this paper, we theoretically show that these four diverse objectives could be derived from a unified theory, \textit{i.e.}, maximizing and minimizing mutual information. In brief, fairness could be achieved by minimizing the mutual information between sensitive attributes $G$ and cluster assignments $C$, \textit{i.e.}, $I(G; C)$; compact and balanced clusters could be obtained by maximizing the conditional mutual information $I(X; C|G)$; and informative features could be learned through maximizing the mutual information $I(X; X^\prime)$, where $X^\prime$ denotes the approximate posterior from the prior. In the following, we will present the mathematical details for fair clustering and informative feature learning in turn.

\subsubsection{Fair Clustering via $\max I(X; C|G)$ and $\min I(G; C)$}
\label{sec:3.2.1}
We begin with the discussion about conditional mutual information $I(X; C|G)$ and mutual information $I(G; C)$. In the following, we will prove why simultaneously maximizing $I(X; C|G)$ and minimizing $I(G; C)$ can make clusters compact, balanced, and fair. First, we formulate the objective function as follows:
\begin{equation}
\label{Eq: I(X;C|G)-gamma I(G; C)}
\begin{aligned}
    \max I(X; C|G) - \gamma I(G; C),
\end{aligned}
\end{equation}
where $\gamma>0$ is a trade-off parameter. By the definition of mutual information, we have
\begin{equation}
\label{eq:3}
\begin{aligned}
    &I(X; C|G) - \gamma I(G; C) \\
    &= H(C|G) - H(C|X,G) - \gamma I(G; C) \\
    &= \left(H(C) - I(G; C) \right) - H(C|X,G) - \gamma I(G; C) \\
    &= - H(C|X,G) + H(C) - (1+\gamma)I(G; C).
\end{aligned}
\end{equation}

Due to the over-high computational complexity, the first term $H(C|X,G)$ is intractable. To overcome this challenge, we theoretically show that $H(C|X,G)$ is exactly $H(C|X)$ which is more computationally efficient. Specifically, as sensitive attributes $G$ are known in prior, we have $H(G|X) = H(G|C, X) = 0$. Hence, $H(C|X,G)$ could be rewritten as:
\begin{equation}
\begin{aligned}
H(C|X, G) &= H(C|X) - I(G; C|X) \\
&= H(C|X) - (H(G|X) - H(G|C,X)) \\
&= H(C|X) = -\frac{1}{N}\sum_{i,k}c_{ik}\log{c_{ik}},
\end{aligned}
\label{Eq: H(C|G,X)}
\end{equation}
where $c_{ik}$ is the probability of sample $x_{i}$ being assigned to the $k$-\textit{th} cluster, which is computed according to its distance to the cluster centers (more details provided later in Supplementary Sec.~2). As can be seen, the minimization of $H(C|X,G)$ pushes each sample to its corresponding cluster center and away from the others, \textit{i.e.}, the model is encouraged to produce \textbf{compact} clusters.

The second term $H(C)$ is the entropy of cluster assignments over all samples, namely,
\begin{equation}
   H(C) = -\sum_{k} p_{\tilde{c}_{k}}\log{p_{\tilde{c}_{k}}},
 \label{Eq. HC}
\end{equation}
where $p_{\tilde{c}_k} = \frac{1}{N}\sum_{i}{c}_{ik}$ refers to the marginal probability density function of clusters. The maximization of $H(C)$ punishes over-large or small clusters to avoid trivial solutions due to over-emphasized fairness, which leads to \textbf{balanced} clusters.

The third term $I(G; C)$ refers to the mutual information between the cluster assignments and the sensitive attributes, which is computed as
\begin{equation}
   I(G; C) = \sum_{t, k}p_{\tilde{g}_t, \tilde{c}_k}\log{\frac{p_{\tilde{g}_t, \tilde{c}_k}}{p_{\tilde{g}_t}p_{\tilde{c}_k}}},
\label{Eq: IGC}
\end{equation}
where $p_{\tilde{g}_t} = \frac{1}{N}\sum_{i} \mathbbm{1}_{x_i \in \tilde{g}_{t}}$ is the group marginal probability density function, and $p_{\tilde{c}_k, \tilde{g}_t} = \frac{1}{N}\sum_{i} \mathbbm{1}_{x_i \in \tilde{g}_{t}}{c}_{ik}$ denotes the joint probability density function of clusters and groups. Notably, $I(G; C)$ is a convex function and it reaches the minimum point $I(G; C) = 0$ \textit{i.f.f.} $p_{\tilde{g}_t, \tilde{c}_k}={p_{\tilde{g}_t}p_{\tilde{c}_k}}$, which exactly corresponds to the absolute fairness defined in Definition~\ref{Def: fair}. Hence, the minimization of $I(G; C)$ encourages the data partition to be fair against sensitive attributes, thus leading to \textbf{fair} clusters. Note that solely optimizing $I(X; C|G)$ would not lead to fair clustering, see Fig.~\ref{Fig:Basic Idea}.

Based on the above analyses, Eq.~\ref{Eq: I(X;C|G)-gamma I(G; C)} could be decomposed into the following two objectives, namely,
\begin{equation}
\begin{aligned}
    \mathcal{L}_{clu} &= -H(C) + H(C|X, G) \\
    &= \sum_{k}p_{\tilde{c}_{k}}\log{p_{\tilde{c}_{k}}} -\frac{1}{N}\sum_{i,k}c_{ik}\log{c_{ik}},
\label{Eq: Lclu}
  \end{aligned}
  \end{equation}
  and
\begin{equation}
\mathcal{L}_{fair} = I(G; C) = \sum_{t,k}p_{\tilde{g}_{t},\tilde{c}_{k}}\log{\frac{p_{\tilde{g}_{t},\tilde{c}_{k}}}{p_{\tilde{g}_{t}}p_{\tilde{c}_{k}}}}.
\label{eq: 3.1.1}
\end{equation}

\subsubsection{Informative Feature Learning via $\max I(X; X^\prime)$}
The above theoretical analysis has shown that compactness, balance, and fairness could be derived from the unified perspective of mutual information. In this section, we reveal that the informative feature could also be learned via maximizing the mutual information $I(X; X^\prime)$ between the prior $X$ and the corresponding approximate posterior $X^\prime$. Without loss of generality, we take $X^\prime$ as the auto-encoder reconstruction for the given $X$ in the following analysis and our implementation.

To be specific, given $I(X; X^\prime) = H(X) - H(X|X^\prime)$, the maximization of $I(X; X^\prime)$ is equivalent to the minimization of the conditional entropy $H(X|X^\prime) = -\mathbb{E}_{p(X,X^\prime)}[\log{p(X|X^\prime)}]$, since $H(X)$ is a constant. However, as the probability density function $p(X|X^\prime)$ cannot be accessed directly, we alternate to minimize its upper bound, \textit{i.e.},
\begin{equation}
\begin{aligned}
&-\mathbb{E}_{p(X,X^\prime)}[\log{q(X|X^\prime)}] \\
&\ge -\mathbb{E}_{p(X,X^\prime)}[\log{q(X|X^\prime)}] - D_{KL}\left(p(X|X^\prime)||q(X|X^\prime)\right) \\
&= -\mathbb{E}_{p(X,X^\prime)}[\log{p(X|X^\prime)}],
\end{aligned}
\end{equation}
where $q(X|X^\prime)$ could be any distribution with a known probability density function and $D_{KL}$ is KL divergence. Without loss of generality, we assume it obeys Gaussian distribution $N(X; X^\prime, \sigma^2 I)$, then
\begin{equation}
\begin{aligned}
-\mathbb{E}_{p(X,X^\prime)}[\log{q(X|X^\prime)}] \propto \mathbb{E}_{p(X,X^\prime)}[||X-X^\prime||^2] + a,
\end{aligned}
\end{equation}
where $a$ is a constant. As a result, the upper bound could be optimized by minimizing the following reconstruction loss, \textit{i.e.},
\begin{equation}
\begin{gathered}
    \mathcal{L}_{rec} = ||X-X^\prime||^{2} = ||X - \Phi(\theta(X))||^2,
\end{gathered}
\label{Eq: L_rec}
\end{equation}
where $\theta$ denotes a shared encoder, and $\Phi$ denotes a multi-branch decoder that reconstructs samples with different attributes separately. Formally, $\Phi(h_{i}) = \phi_{g_{i}}(h_{i}), g_{i} \in [1, 2, \dots, T]$, where $\phi_{g_{i}}$ is the group-specific decoder which reconstructs samples from the $g_{i}$-\textit{th} group, and $h_{i} = \theta(x_{i})$ denotes the feature extracted by the encoder. In other words, we use a multi-branch decoder to recover the group information that has been removed in the hidden space by $\mathcal{L}_{\text{fair}}$ for better reconstruction.

As $I(X; X^\prime) \leq I(X; h)$, the maximization of $I(X; X^\prime)$ intrinsically increases the lower bound of $I(X; h)$, and thus helps the auto-encoder to extract \textbf{informative} features $h$ from the raw inputs.

\subsection{The Objective Function and Algorithm Details}
\label{Sec:OverallObectiveFunction}
To summarize, we unify the deep fair clustering task from the perspective of information theory. With the above theoretical analyses, we arrive at our loss by combining Eq.~\ref{Eq: Lclu},~\ref{eq: 3.1.1} and~\ref{Eq: L_rec}, \textit{i.e.},
\begin{equation}
\begin{aligned}
    \mathcal{L} &= \mathcal{L}_{rec} + \alpha (\mathcal{L}_{clu} + (1+\gamma)\mathcal{L}_{fair})\\
    &= \mathcal{L}_{rec} + \alpha \mathcal{L}_{clu} + \beta \mathcal{L}_{fair},
\end{aligned}
\label{Eq: Objective Function}
\end{equation}


\subsection{The Proposed Evaluation Metrics for Fair Clustering}
To evaluate the fairness of clustering results, most existing studies adopt the \textit{Balance (Bal.)} metric~\cite{fairlets,Towards} which is defined as the ratio between the largest and smallest sensitive groups in a cluster.
However, the distributions of other groups are ignored when there are more than two groups (e.g., if there were 3, 4, 18, 20 samples from each group, the \textit{Bal.} will be 3/20 which is the same when it becomes 3, 11, 11, 20.) To address this issue, we propose a novel fairness measurement, dubbed~\textit{Minimal Normalized Conditional Entropy (MNCE)} as below. 

\begin{definition}
    \textbf{Minimal Normalized Conditional Entropy (MNCE).} Given $N$ data points $X$ with sensitive groups $G$ from clusters $C$, MNCE is defined as the minimal group entropy in each cluster divided by the global group entropy. Formally,
\begin{equation}
\begin{aligned}
    \mathrm{MNCE} &= \frac{\underset{k}{min}\left(H\left({G|\tilde{c}_{k}}\right)\right)}{H(G)} \\
    &= \frac{\underset{k}{min}\left(-\sum_{t} \frac{|\tilde{g}_{t} \cap \tilde{c}_{k}|}{|\tilde{c}_{k}|}\log{\frac{|\tilde{g}_{t} \cap \tilde{c}_{k}|}{|\tilde{c}_{k}|}}\right)}{-\sum_{t} \frac{|\tilde{g}_{t}|}{N}\log{\frac{|\tilde{g}_{t}|}{N}}} \in [0,1],
\label{Eq: MNCE}
\end{aligned}
\end{equation}
where $\tilde{g}_t=\{x_i|g_i=t\}$ and $\tilde{c}_k=\{x_i|c_i=k\}$ denote the set of samples belonging to the $t$-\textit{th} group and $k$-\textit{th} cluster respectively, $H(G|\tilde{c}_{k})$ is the conditional entropy of sensitive attributes given the cluster assignments, and $H(G)$ denotes the entropy of sensitive attributes.
\end{definition}

As $H(G)$ is a constant for the given dataset, we derive the following proposition from Eq.~\ref{Eq: MNCE}:
\begin{proposition}
\label{proposition: mnce}
The cluster assignments are independent of the sensitive attributes if and only if $\mathrm{MNCE} = 1$.
\end{proposition}
\begin{proof}
Suppose the cluster assignments are independent of the sensitive attributes, we have ${p_{\tilde{g}_t,\tilde{c}_k}}/{p_{\tilde{c}_k}} = p_{\tilde{g}_t}$, where $p_{\tilde{c}_k, \tilde{g}_t}$ denotes the joint probability density function of clusters and groups,  $p_{\tilde{g}_t}$ and $p_{\tilde{c}_k}$ are the marginal probability density function of group and clusters, respectively. From the Bayes' theorem, one have $p_{g_t|c_k} = {p_{g_t,c_k}}/{p_{c_k}}$, and thus $p_{g_t|c_k} = p_{g_t}$. Further,
\begin{equation}
\begin{aligned}
    H({G|\tilde{c}_k}) &= - \sum_{t} p_{\tilde{g}_t|\tilde{c}_k}\log{p_{\tilde{g}_t|\tilde{c}_k}} = - \sum_{t} p_{\tilde{g}_t}\log{p_{\tilde{g}_t}} \\
    &= H(G), \forall k \in \{1, 2, \dots, K\}.
\end{aligned}
\end{equation}
As a result, $\underset{k}{min}(H({G|\tilde{c}_k})) = H(G)$ and $\mathrm{MNCE} = 1$. Notably, the above deductions hold reversely as $H({G|\tilde{c}_k}) \leq H(G), \forall k$.
\end{proof}

From Proposition~\ref{proposition: mnce}, \textit{MNCE} reaches the maximum point \textit{MNCE} $= 1$ \textit{i.f.f.} the cluster assignments are absolutely \textit{fair} as defined in Definition~\ref{Def: fair}. Hence, a larger \textit{MNCE} indicates a fairer clustering result.

However, a separate evaluation is less attractive as fair clustering methods usually make a trade-off between clustering quality and fairness. Alternatively, we propose the following metric to simultaneously measure the clustering quality and fairness, 
\begin{definition}
\textbf{$F_{\beta}$: An overall measure for clustering quality and fairness.} Let $u$ and $v$ denote the clustering metric NMI and the fairness metric MNCE, $F_{\beta}$ is defined as the harmonic mean of $u$ and $v$, \textit{i.e.},
\begin{equation}
\begin{aligned}
    F_{\beta} &= 1/\left(\frac{\beta^{2}}{1+\beta^{2}}\frac{1}{v}+\frac{1}{1+\beta^{2}}\frac{1}{u}\right) \\
    &= \frac{(1+\beta^{2})uv}{\beta^{2}u + v} \in [0,1] ~ (u, v \in [0,1]),
\end{aligned}
\label{Eq: Fbeta}
\end{equation}
where $\beta \in [0, +\infty)$ is a hyper-parameter to adjust the weight of clustering quality and fairness. A larger $\beta$ corresponds to more focus on fairness. In general, we recommend $\beta=1$ to treat these two terms equally.
\end{definition}

\section{Experiments}
\label{Sec.Experiments}
\begin{table*}
  \caption{Comparisons of FCMI with both standard and fair clustering methods on six benchmarks. The best and the second best results are marked in \textbf{bold} and \underline{underline}, respectively. Some results are unavailable since the codes for DFDC and Toward are unpublished, and DFC and ScFC are unable to handle the HAR dataset which is consisted of multiple ($\ge 3$) sensitive groups.}
  \label{Tab.MainTable1}
  \centering
\resizebox{\textwidth}{!}{%
\setlength{\tabcolsep}{3.75pt}
      \begin{tabular}{lccccccccccccccc}
      \toprule
      &\multicolumn{5}{c}{MNIST-USPS}&\multicolumn{5}{c}{Color Reverse MNIST}&\multicolumn{5}{c}{HAR}\\
      \midrule
      Method & ACC & NMI & Bal & MNCE & $F_{\beta}$  & ACC & NMI & Bal & MNCE &  $F_{\beta}$  & ACC & NMI & Bal & MNCE &  $F_{\beta}$ \\
      \midrule
      AE   & 76.3 & 71.8 & 0.0 & 0.0 & 0.0 & 41.0 & 52.8 & 0.0 & 0.0 & 0.0 & 66.3 & 60.7 & 0.0 & 86.9 & 71.5 \\
      DEC   & 60.0 & 59.4 & 0.0 & 0.0 & 0.0 & 40.7 & 38.2 & 0.0 & 0.0 & 0.0 & 57.1 & 65.5 & 0.0 & 93.7 & 77.1 \\
      DAC    & 76.3 & 69.9 & 0.0 & 0.0 & 0.0 & 31.4 & 27.1 & 0.0 & 0.0 & 0.0 & 38.2 & 31.5 & 0.0 & 32.4 & 31.9 \\
      ClGAN  & 38.3 & 35.7 & 0.1 & 1.9 & 3.6 & 20.1 & 9.1 & 2.2 & 14.9 & 11.3 & 52.7 & 44.3 & 0.4 & 0.0 & 0.0 \\
      \midrule
      ScFC   & 14.2 & 1.3 & \underline{11.2} & \textbf{95.0} & 2.6 & 51.3 & 49.1 & \textbf{100.0} & \textbf{100.0} & 65.8 & - & - & - & - & - \\
      SpFC  & 20.1 & 15.5 & 0.0 & 0.0 & 0.0 & 11.0 & 2.1 & 0.0 & 0.0 & 0.0 & 19.0 & 0.4 & 0.0 & 0.0 & 0.0  \\
      VFC   & 58.1 & 55.2 & 0.0 & 0.0 & 0.0 & 38.1 & 42.7 & 0.0 & 0.0 & 0.0 & 62.6 & 66.2 & 25.6 & 98.7 & \underline{79.3} \\
      FAlg   & 58.4 & 53.8 & 9.5 & 85.8 & 66.1 & 26.9 & 14.3 & 66.6 & 97.1 & 24.9 & 56.6 & 58.6 & \underline{43.2} & \underline{99.2} & 73.7 \\
      Towards   & 72.5 & 71.6 & 3.9 & - & - & 42.5 & 50.6 & 43.0 & - & - & 60.7 & 66.1 & 16.6 & - & - \\
      DFC & 85.7 & 83.4 &  6.7&68.2  & \underline{75.0} &  49.9                & 68.9 &80.0  & 99.1 & \underline{81.3} & - & - & - & - & -  \\
      DFDC &  \underline{93.6}  &  \underline{87.6}  &  \textbf{11.9}        &-  &-      &  \underline{58.9}     &  \underline{69.0}  &  {94.6}    &-  &-      &  \underline{86.2}       &  \textbf{84.5}   &  \textbf{46.8}     &- &-\\
      FCMI(Ours) & \textbf{96.7} & \textbf{91.8} &  10.7 & \underline{94.5} & \textbf{92.0} & \textbf{88.4} & \textbf{86.4} & \underline{99.5} &  \underline{99.9} & \textbf{92.7} & \textbf{88.2} &\underline{80.7} & 40.7 & \textbf{99.3} & \textbf{89.0} \\
      \toprule
      &\multicolumn{5}{c}{Office-31}&\multicolumn{5}{c}{MTFL}&\multicolumn{5}{c}{Mouse Atlas}\\
      \midrule
      Method & ACC & NMI & Bal & MNCE &  $F_{\beta}$  & ACC & NMI & Bal & MNCE &  $F_{\beta}$  & ACC & NMI & Bal & MNCE &  $F_{\beta}$ \\
      \midrule
      AE   & 63.8 & 66.8 & 0.0 & 0.0 & 0.0 & 67.2 & 16.0 & 67.8 & 97.3 & 27.5 & 56.1 & 54.5 & 0.6 & 5.7 & 10.3 \\
      DEC   & 63.3 & 68.6 & 0.0 & 0.0 & 0.0 & 56.7 & 0.6 & 78.0 & 98.9 & 1.1 & \underline{61.5} & \underline{63.2} & 0.7 & 6.3 & 11.4 \\
      DAC   & 14.0 & 25.2 & 0.0 & 0.0 & 0.0 & 58.9 & 1.4 & 81.5 & 87.9 & 2.7 & 48.3 & 40.3 & 36.0 & 86.4 & 54.9 \\
      ClGAN   & 52.2 & 54.9 & 0.0 & 0.0 & 0.0 & \textbf{72.9}  & 12.6 & 79.1 & 99.0 & 22.4 & 48.3 & 50.8 & 0.8 & 7.2 & 12.6 \\
      \midrule
      ScFC   & 38.0 & 60.7 & \textbf{26.7} & \textbf{97.7} & 74.9 & 52.1 & 15.1 &\textbf{100.0} & \textbf{100.0} & 26.3 & 31.2 & 16.0 & \textbf{63.6} & \textbf{99.9} & 27.5 \\
      SpFC  & 9.3 & 11.4 & 0.0 & 0.0 & 0.0 & 65.5 & 0.1 & 75.0 & 98.5 & 0.2 & 21.3 & 3.8 & 0.0 & 0.0 & 0.0 \\
      VFC   & 65.2 & 69.7 & 20.3 & 86.0 & 77.0 & 68.8 & 8.4 & 88.9 & 99.8 & 15.6 & 45.4 & 49.8 & 0.0 & 0.0 & 0.0 \\
      FAlg    & 67.1 & 70.7 & 20.4 & 86.4 & \underline{77.8}& 63.2 & 16.7 & 60.1 & 96.3 & 28.5 & 52.2 & 58.5 & \underline{45.2} &\underline{92.7}  &\underline{71.7} \\
      DFC   & \underline{69.0} & \underline{70.9} & 11.9 & 64.2 & 67.4 & \underline{72.8} & \underline{17.6} &\underline{97.4}& \underline{99.9} & \underline{30.0}& 60.0 & 59.1 & 21.4 & 69.7 & 63.9  \\
      FCMI(Ours)  & \textbf{70.0} & \textbf{71.2} & \underline{22.6} & \underline{90.6} & \textbf{79.7} & 70.2 & \textbf{19.1} &  90.4 & 99.8 & \textbf{32.0} & \textbf{ 65.8 } & \textbf{65.4} & 38.1 & {88.1} & \textbf{75.0} \\
      \bottomrule
  \end{tabular}
}
\end{table*}
\begin{figure*}[t]
\centering
  \includegraphics[width=\textwidth]{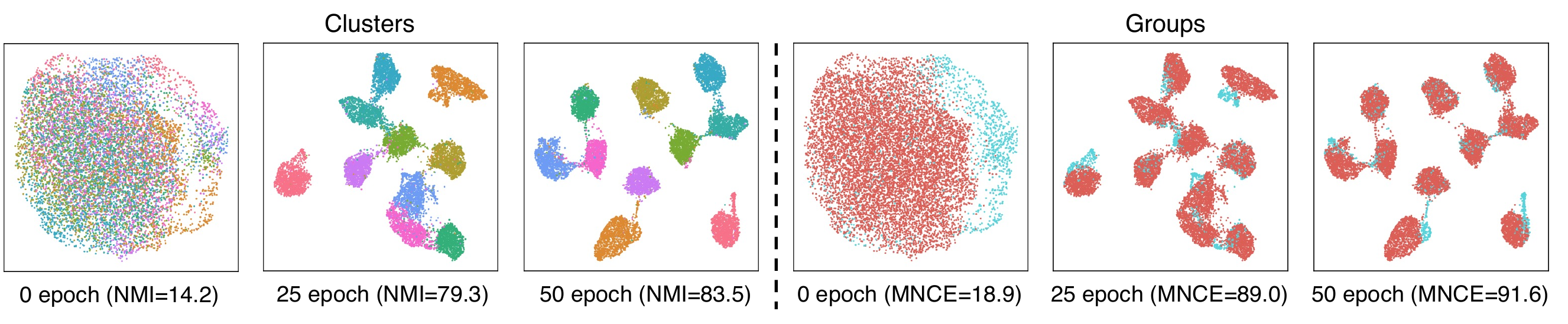}
\caption{Visualization of the hidden representation on MNIST-USPS with the increasing training epoch. The left and right three figures are colored by classes and groups, respectively. For more visualization and comparison, please see Supplementary Fig~1.}
\label{Fig: VisualFeature}
\end{figure*}
In this section, we evaluate our FCMI on six fair clustering benchmarks, compared with both the classical (non-fair) and state-of-the-art fair clustering methods. In addition, we carry out a series of qualitative analyses and ablation studies to investigate the effectiveness and robustness of FCMI.

\subsection{Experimental Setups}
\label{sec: setup}

\textbf{Dataset:} We adopt six datasets confounded with various types of sensitive attributes for evaluations (summarized in Supplementary Table~1). Among them, the first five datasets including MNIST-USPS\footnote{\url{http://yann.lecun.com/exdb/mnist},~\url{https://www.kaggle.com/bistaumanga/usps-dataset}}, Reverse MNIST, HAR~\cite{anguita2013public}, Office-31~\cite{saenko2010adapting}, and MTFL~\cite{zhang2014facial} are commonly used in fair clustering studies~\cite{Towards, DFCV}. To explore the potential in practical applications, we additionally evaluate FCMI on the single-cell mouse atlas dataset that is widely used in biological analysis, where the sensitive attributes correspond to different sequence techniques~\cite{MouseCellAtlas1, MouseCellAtlas2}.

\textbf{Implementation Details:}
Consistent with the previous works~\cite{DFCV}, we use a convolutional auto-encoder for MNIST-USPS and Reverse MNIST, and a fully-connected auto-encoder for handing other datasets. For MTFL and Office-31, the features extracted by ResNet50~\cite{he2016deep} are used as the inputs. In all experiments by default, we fix the hyper-parameter $\alpha = 0.04, \beta = 0.20$ in Eq.~\ref{Eq: Objective Function} across all the datasets. The only exception is that we remove the balance constraint (\textit{i.e.}, set $\beta = 0$) on the single-cell mouse atlas since the cells of different types are highly unbalanced. The model is trained for 300 epochs using the Adam optimizer with an initial learning rate of $1e-4$ for all datasets, with a warm-up in the first 20 epochs using the reconstruction loss defined in Eq.~\ref{Eq: L_rec}. All experiments are conducted on a Nvidia A10 GPU on the Ubuntu 18.04 platform.

\textbf{Baselines:}
Both classic clustering methods and state-of-the-art fair clustering methods are used for comparisons. Specifically, for the classic methods, we select auto-encoder + k-means~\cite{AE}, DEC~\cite{DEC}, DAC~\cite{DAC} and ClGAN~\cite{ClGAN} as baselines. For the shallow fair clustering methods, ScFC~\cite{Scalable_fair_clustering}, SpFC~\cite{SpFC}, VFC~\cite{FCC}, FAlg~\cite{FALG} are used for comparisons. For the deep fair clustering methods, we could only investigate the performance of DFC~\cite{DFCV} since the code of DFDC~\cite{DFDC} and Towards~\cite{Towards} are unavailable. As an alternative, we present the original results reported in their paper on the common datasets for reference. Notably, ScFC~\cite{Scalable_fair_clustering} and DFC~\cite{DFCV} only support two groups and they are impractical on the HAR dataset, while other methods including our FCMI could be generalized to arbitrary group numbers.

\textbf{Evaluation Metrics:}
In our experiments, the widely-used \textit{ACC} and \textit{NMI} metrics are used to investigate the clustering quality. And the previous \textit{Balance (Bal.)} metric and the proposed \textit{MNCE} metric are used to evaluate the fairness. In addition, we adopt the proposed measurement $F_{\beta}$ for a comprehensive evaluation.

\subsection{Quantitative Comparisons}

In this section, we carry out quantitative experiments by comparing FCMI with 11 baselines. As shown in Tab.~\ref{Tab.MainTable1}, although the classical clustering methods achieve competitive clustering performance in terms of \textit{ACC} and \textit{NMI}, they show poor results in terms of fairness. On the contrary, fair clustering methods inject fairness into clustering, leading to a debiased data partition. However, some shallow fair clustering methods such as ScFc~\cite{Scalable_fair_clustering} guarantee the fairness explicitly at the cost of clustering quality. Compared with shallow ones, deep fair clustering methods achieve a more elegant trade-off between clustering quality and group fairness. Besides the superior performance of our FCMI on clustering and fairness metrics, we would like to highlight that FCMI achieves dominance in terms of $F_\beta$. Specifically, FCMI outperforms the best competitor by $17.0\%$, $11.4\%$, $9.7\%$, $1.9\%$, $2.0\%$, and $3.3\%$ in terms of $F_\beta$ on six datasets respectively. 

\subsection{Visualization}
To help understand the working mechanism of FCMI, we first visualize the hidden representation of MNIST-USPS by performing UMAP~\cite{mcinnes2018umap} on the learned features across the training process. As shown in Fig.~\ref{Fig: VisualFeature}, the data shows the group information evenly distributes in the initial feature space. As the training goes on, more compact, balanced, and fairer clusters are learned by FCMI.

\begin{figure}[t]
  \centering
  \includegraphics[width=\linewidth]{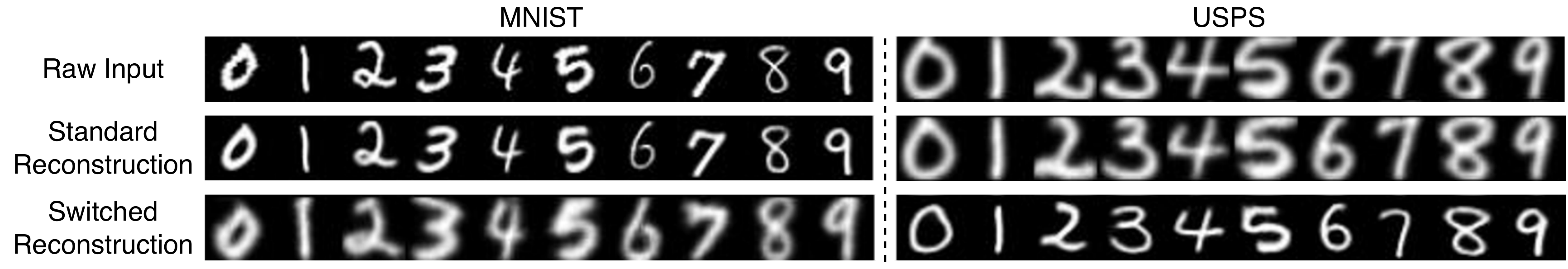}
  \caption{Visualization of the reconstructions outputted by group-specific decoders on MNIST-USPS. The rows from top to bottom refer to the raw inputs, the reconstructions from the corresponding decoder, and the reconstructions from the switched decoder.}
  \label{Fig: VisualReconstruction}
\end{figure}
Recalling in our implementation, a multi-branch decoder is used to recover the group information for better reconstruction. To verify the effectiveness of such a multi-branch decoder, we conduct experiments on MNIST-USPS and switch the decoder for different groups to reconstruct images. As shown in Fig.~\ref{Fig: VisualReconstruction}, the group information is successfully transferred from USPS to MNIST,  and vice versa. In other words, this result proves that the encoder could extract the semantic information and each branch of the decoder could  capture the group information.

\subsection{Parameter Analysis and Ablation Study}
\label{Sec.Ablation}    

\begin{figure}
\centering
\includegraphics[width=\linewidth]{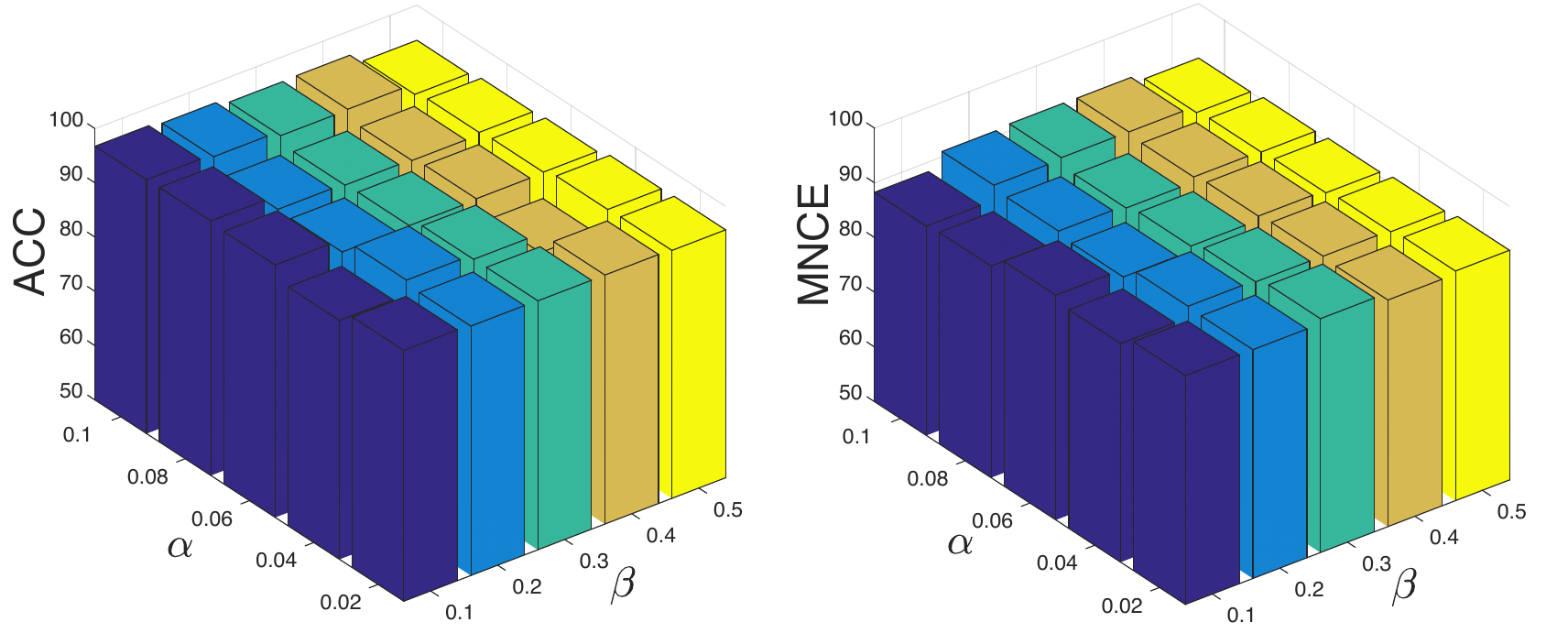}
\caption{The influence of $\alpha$ and $\beta$ on \textit{ACC} and \textit{MNCE} on MNIST-USPS. The stable performance of FCMI demonstrates its robustness against hyper-parameters.}
\label{Fig: AblationParameterAnalysis}
\end{figure}
\begin{table}
\captionof{table}{Ablation study of $\mathcal{L}_{clu}$ and $\mathcal{L}_{fair}$ on MNIST-USPS.}
\centering
   \begin{tabular}{cc|ccccc}
       \toprule
       $\mathcal{L}_{clu}$ & $\mathcal{L}_{fair}$ & ACC  & NMI & Bal & MNCE & $F_{\beta}$ \\
       \midrule
        $\checkmark$ & $\checkmark$ & \textbf{96.7}  &  \textbf{91.8}  & \textbf{10.7} & \textbf{92.3} & \textbf{92.0} \\
        $\checkmark$ & & \underline{95.2}  &  \underline{89.9}  &  8.2 & 77.9 & \underline{83.5} \\
        & $\checkmark$ & 89.6  &  79.2  &  \underline{9.0} & \underline{86.1} & 81.1 \\
       \bottomrule
   \end{tabular}
\label{tab:ablation}
\end{table}


In this section, we investigate the influence of the hyper-parameters $\alpha$ and $\beta$ on the MNIST-USPS dataset. As shown in Fig.~\ref{Fig: AblationParameterAnalysis}, the performance of FCMI is stable under different choices of $\alpha$ and $\beta$, which demonstrates its robustness against the hyper-parameters. However, when one of the mutual information $I(X; C|G)$ and $I(G; C)$ is removed (\textit{i.e.}, without $\mathcal{L}_{clu}$ or $\mathcal{L}_{fair}$), our model encounters a significant drop in clustering or fairness performance as shown in Table~\ref{tab:ablation}. Such an ablation study verifies the effectiveness of our information theory-driven losses.

\begin{figure}
\centering
\includegraphics[width=\linewidth]{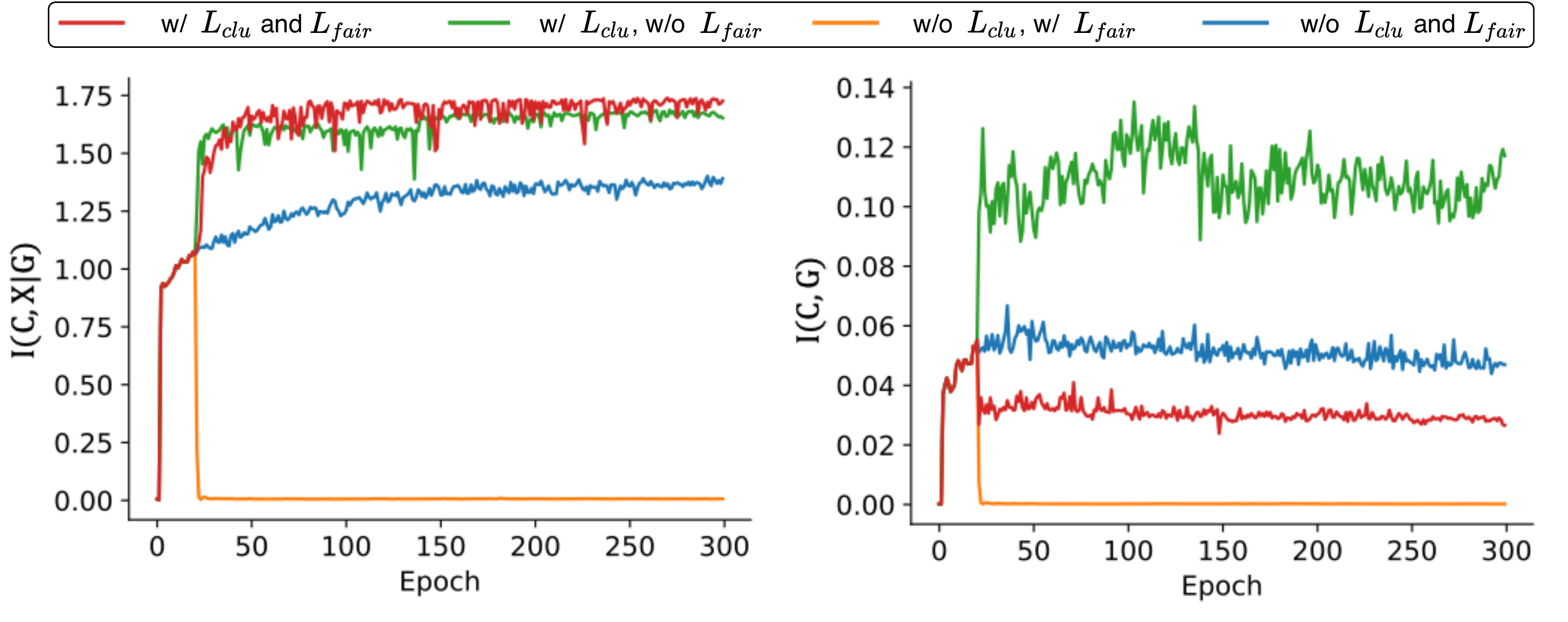}
  \caption{The evolution of mutual information $I(C, X|G)$ and $I(C, G)$ across the training process with and without $\mathcal{L}_{clu}$ and $\mathcal{L}_{fair}$.}
  \label{Fig: AblationObjectiveVsEpoch}
\end{figure}

To further investigate the effectiveness of $\mathcal{L}_{clu}$ and $\mathcal{L}_{fair}$, we visualize the evolution of mutual information $I(X; C|G)$ and $I(G; C)$ across the training process on the HAR dataset. For comparisons, we compute the mutual information when we remove one or both losses. As demonstrated in Fig.~\ref{Fig: AblationObjectiveVsEpoch}, both $I(X; C|G)$ and $I(G; C)$ increase in the first $20$ epochs since the informative features contain the cluster and group information at the preliminary learning stage. After that, with $\mathcal{L}_{clu}$ and $\mathcal{L}_{fair}$ (red line), the model becomes more powerful to differentiate different clusters while alleviating the influence of the group information, compared with the baseline when only $\mathcal{L}_{rec}$ is used (blue line). Without $\mathcal{L}_{fair}$ (green line), group information will leak into the cluster assignments resulting in unfair data partitions. Without $\mathcal{L}_{clu}$ (orange line), the model would collapse by solely minimizing mutual information $I(G; C)$. 

\section{Conclusion}
\label{Sec.Conclusion}
In this paper, we build a novel deep fair clustering method (FCMI) and theoretically show that it could achieve compact, balanced, and fair clusters, as well as informative features. In addition, we design a novel evaluation metric that measures the clustering quality and fairness as a whole. Extensive experimental results demonstrate the superiority of our method over 11 baselines on six benchmarks including a single-cell RNA-seq atlas.
\section*{Acknowledgement}
This work was supported in part by the National Key R\&D Program of China under Grant 2020YFB1406702; in part by NSFC under Grant U21B2040, 62176171, and U19A2078; in part by Sichuan Science and Technology Planning Project under Grant 2022YFQ0014.

{\small
\bibliographystyle{ieee_fullname}
\bibliography{reference}
}

\end{document}